\documentclass{article}

\usepackage{arxiv}
\usepackage{custom-setup}

\title{Finetuning Vision-Language Models as OCR Systems for Low-Resource Languages: A Case Study of Manchu}

\author{%
  Yan Hon Michael Chung\\
  Division of Humanities\\
  The Hong Kong University of Science and Technology\\
  Hong Kong SAR\\
  \texttt{hmichung@ust.hk}\\
  \And~Donghyeok Choi\\
  History, Religious and Philosophy Academy\\
  Hong Kong Baptist University\\
  Hong Kong SAR\\
  \texttt{dhchoi@hkbu.edu.hk}%
}

\begin{document}
\maketitle

\begin{abstract}
	Manchu, a critically endangered language essential for understanding early modern Eastern Eurasian history, lacks effective OCR systems that can handle real-world historical documents. This study develops high-performing OCR systems by fine-tuning three open-source vision-language models (LLaMA-3.2-11B, Qwen2.5-VL-7B, Qwen2.5-VL-3B) on 60,000 synthetic Manchu word images using parameter-efficient training. LLaMA-3.2-11B achieved exceptional performance with 98.3\% word accuracy and 0.0024 character error rate on synthetic data, while crucially maintaining 93.1\% accuracy on real-world handwritten documents. Comparative evaluation reveals substantial advantages over traditional approaches: while a CRNN baseline achieved 99.8\% synthetic accuracy, it suffered severe degradation to 72.5\% on real documents. Our approach demonstrates effective synthetic-to-real domain transfer, providing a cost-effective solution deployable on accessible infrastructure. This work establishes a transferable framework for endangered language OCR that removes technical and financial barriers in digital humanities, enabling historians and linguists to process historical archives without specialized computing resources.\footnote{Code and model weights are available at \url{https://github.com/mic7ch1/ManchuAI-OCR}}
\end{abstract}


\section{Introduction}
\label{sec:introduction} In 1993, Pamela Kyle Crossley and Evelyn S. Rawski published a seminal article in the \textit{ Harvard Journal of Asiatic Studies} titled ``A Profile of the Manchu Language in Ch'ing History'', in which they argued that the Manchu language and its archival sources are essential for a comprehensive understanding of Qing political, administrative, cultural, and frontier histories. The Qing, the last imperial dynasty of China, was ruled by the Manchus, an ethnic minority whose now critically endangered native language, Manchu, served as both a court and administrative language of the empire.  Challenging the long-standing assumption that Manchu-language materials merely duplicated their Chinese counterparts, Crossley and Rawski demonstrated that many Manchu documents contained unique content—particularly in sensitive political, military, and ideological contexts~\citep[p.~102]{crossley1993}. With growing access to Manchu archival materials, they anticipated that ``the relevance of the language (Manchu) to any particular category of research (of Qing history) is not as predictable as it was once assumed to be''. This call for a methodological shift was reinforced in~\citeauthor{Rawski_1996}'s 1996 presidential address to the Association for Asian Studies, where she highlighted how a Manchu-centered perspective fundamentally altered prevailing narratives of the Qing.~\citep[pp.~829--830]{Rawski_1996} Together, these interventions catalyzed the emergence of what is now known as the ``New Qing History'', a scholarly approach that repositions the Qing as a multiethnic empire rather than a purely sinicized Chinese dynasty.

That said, the language barrier has remained a major obstacle to the broader utilization of the vast Manchu archival corpus. Proficiency in Manchu remains limited to a small circle of specialists, and only a fraction of the available sources has been systematically catalogued or studied. Recent advances in Vision Language Models (VLMs) offer new possibilities for expanding access to these materials. This study introduces a high-performing OCR system for the Manchu language, developed by fine-tuning the Llama-3.2-11B-Vision model on a synthetically generated dataset comprising 60,000 Manchu word images. The resulting system achieves a remarkably low character error rate (CER) of 0.0024 and a word-level accuracy of 98.3\% on synthetic data, while effectively maintaining a CER of 0.0219 and word-level accuracy of 93.1\% on real-world handwritten Manchu word images. 

This proposed Manchu OCR system not only facilitates the reading of Manchu-language texts but also helps promote their wider use in historical research. Additionally, this study presents a simple and transferable framework which requires only modest synthetic data and computing resources for building robust OCR systems for low-resource and endangered languages. The recent advances in VLM AI thus open promising opportunities for the preservation and revitalization of many other historically significant, low-resource, and endangered languages that remain underserved by existing OCR technologies.
\section{Literature Review}
\label{sec:related_work}

\subsection{Earlier OCR Works}

Efforts to develop an Optical Character Recognition (OCR) system for the Manchu language began in the mid-2000s. Early research focused on recognition of primitive Manchu characters at the character level. They proposed a stroke-based method that decomposed handwritten Manchu characters into basic stroke elements, matched these to a predefined stroke library and reassembled them into Manchu-Roman code using fuzzy string matching~\citep{zhang_2004offline}. However, this approach was constrained by the complexity of stroke combination rules and the high variability of handwriting styles, which led to recognition errors in cases of distorted or noisy inputs. Subsequent efforts continued along the character-level trajectory, developing recognition systems that extracted and classified Manchu Character Units (MCUs), a sub-character structural unit~\citep[pp.~802-803]{zhao_2006design}~\citep[pp.~3339-3340]{zhang_2006}. While these models introduced some advances, they still relied on manual segmentation and struggled with compounded character forms and the lack of spacing between letters in the vertical Manchu script.

A significant methodological shift occurred around 2017 with the adoption of deep neural network architectures, particularly convolutional neural networks (CNNs). Researchers began to move away from character-level segmentation and instead adopted word-level, segmentation-free recognition frameworks.~\citep[pp.~46-49]{huang_2017} Building on this direction, researchers further improved word-level recognition accuracy by training models on large synthetic Manchu word datasets, generated through augmentation techniques such as rotation, blur, and elastic distortion. These datasets were derived from 2,135 word images representing 666 unique Manchu words extracted from the textbook 365 Sentences in Manchu, with each word class expanded into 1,000 variants to create a corpus of 666,000 samples. These CNN models could achieve a peak accuracy of 97.68\% and an average accuracy above 91\% in printed fonts.~\citep{Li_2018, Zheng_2018} A subsequent study conducted in 2021 adopted a similar segmentation-free approach using a deep CNN architecture and a sliding window mechanism. Their system was trained on a slightly expanded dataset of 671 classes, also with 1,000 samples per class, and achieved a peak accuracy of 98.84\%.~\citep[pp.~5-6]{Zhang_2021} However, it is important to note that all these models were trained and evaluated on variations of a closed set of fewer than 700 Manchu words, and their performance on unseen vocabulary remains untested. Consequently, while these studies demonstrate strong performance in controlled environments, their generalizability to the full range of historical Manchu texts has yet to be established.

Beyond academic research, several institutions and individuals have also developed practical OCR systems for Manchu. The First Historical Archives of China in Beijing has reportedly created an OCR system tailored for scanned handwritten Manchu documents. However, as the system is restricted to internal institutional use, its technical specifications and recognition accuracy remain unavailable for public evaluation. In 2022, Zhang Zhuohui released an open-source OCR tool for Manchu developed in Python on GitHub~\citep{tyotakuki2022manchuocr}. While the system demonstrates promising functionality in processing printed Manchu texts, no formal assessment of its recognition accuracy has yet been published.

More recent research has explored the integration of linguistic understanding and visual features to overcome the limitations of purely visual OCR approaches in low-resource scripts. The most advanced contribution to date is the work of \citeauthor{Wang_2024}'s team, which introduced the Visual-Language framework for Manchu word Recognition (VLMR). The model fuses visual features with semantic reasoning carried out by a language module. These two streams are linked by a self-teacher network, which jointly refines visual and linguistic representations through self-knowledge distillation. Trained on 21,245 woodblock printed and 6,721 handwritten Manchu word samples, VLMR achieved 98.24\% character recognition accuracy (CRA) and 93.84\% word recognition accuracy (WRA) on the Woodblock Manchu Word (WMW) dataset. In the more challenging Handwritten Manchu Word (HMW) dataset, which includes 6,721 samples, it achieved 95.10\% CRA and 84.08\% WRA~\citep{Wang_2024}. These results demonstrate that integrating visual and semantic cues—especially in a multimodal learning framework—can yield state-of-the-art performance even with constrained training data.

Complementing this direction, recent experiments have begun to fine-tune general-purpose VLMs for OCR tasks in similarly underserved linguistic contexts. A notable case is Miguel Escobar Varela's adaptation of Qwen2-VL-2B-Instruct model for recognizing historical Malay texts in the Jawi script, an Arabic-derived orthography characterized by cursive ligatures, letter ambiguity, and sparse digital resources. Despite training on just 911 annotated samples, the model achieved a character error rate (CER) of 8.66\% and a word error rate (WER) of 25.50\%.~\citep{qwen-for-jawi-v1} This pioneering result demonstrates that even compact, generalist VLM can be effectively fine-tuned for specialized OCR tasks, offering scalable and functional solutions for historically significant but digitally marginalized scripts.

\subsection{Manchu Script and Romanization}
\begin{center}
	\includegraphics[width=0.45\columnwidth]{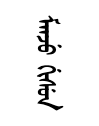}
	\captionof{figure}{manju gisun, meaning `Manchu language'.}
	\label{fig:manju_gisun}
\end{center}
The Manchu script is generally regarded as an alphabetic syllabary, derived from the vertically written Mongolian script, which in turn comes from Old Uyghur. It consists of individual letters representing both vowels and consonants. These letters vary in shape depending on their position within a syllable—initial, medial, or final—and are visually connected in flowing vertical strokes. While not joined into complete words as in Latin cursive, the syllables are linked vertically, giving the appearance of uninterrupted script. Over time, diacritics and modified letter forms were introduced to improve phonetic clarity and standardization.

To transliterate the Manchu script into Latin script, scholars have developed a variety of romanization systems, each reflecting different phonological emphases and typographic conventions. This study adopts the modified M{\"o}llendorff system as presented by Roth Li, which strikes a balance between phonetic precision and typographic simplicity~\citep[p.~16]{li2000manchu}. Diacritics are avoided when possible, and Latin letters are selected to closely approximate Manchu phonemes while maintaining accessibility for learners.

\section{Dataset}
This study adopts the open-source synthetic Manchu dataset developed by Zhang Zhuohui, with substantial additional processing and formatting to prepare it for vision-language model training~\citep{tyotakuki2022manchuocr}. Zhang's dataset was created through a modular pipeline that systematically generates and preprocesses word-level image samples for OCR training. The foundation of the dataset is a curated comprehensive dictionary of 130,917 Manchu words. From this lexicon, individual words were randomly sampled and rendered as images using multiple distinct Manchu typefaces. Each font contains a set of pre-rendered grayscale images corresponding to Manchu words. These images were retrieved and paired with their textual labels to ensure accurate alignment between visual input and semantic content. A total of 750,000 training samples were synthesized, along with an evaluation set of 25,000 samples. 

For the present study, a random subset of 60,000 word images from the full dataset was selected and subjected to additional preprocessing tailored to the requirements of VLM architecture. Specifically, the images, originally rendered as white glyphs on a black background with synthetic noise, were color-inverted to a black-on-white format, denoised, and contrast-enhanced to improve clarity and reduce background artifacts. All images were also resized to fixed dimensions, ensuring uniformity across the dataset. As shown in Figures~\ref{fig:comparision_original_preprocessed}, this transformation produced cleaner, higher-contrast inputs, facilitating more effective model learning and improving OCR accuracy.


\begin{center}
	\includegraphics[width=1\columnwidth,clip,trim=0.1cm 0.1cm 9cm 0.1cm]{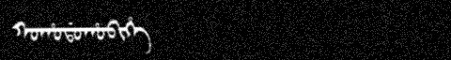}
	\includegraphics[width=1\columnwidth,clip,trim=0.1cm 0.1cm 9cm 0.1cm]{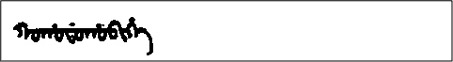}
	\captionof{figure}{Original and preprocessed Manchu word images}
	\label{fig:comparision_original_preprocessed}
\end{center}

The evaluation dataset for the current study consists of two parts. First, 15,000 synthetic word images were used as a validation set during training, evaluated every 1,000 steps to monitor the model's performance. For final model performance assessment, we randomly sampled 1,000 Manchu word images from the synthetic dataset and measured each model's evaluation metric. Second, a real-world test set was constructed from 218 handwritten Manchu word images, extracted from the scanned Manchu document \textit{Neige Cangben Manwen Laodang}, a compilation of historical documents created by the Qing court in the 18th century~\citep[p.~3]{neige_cangben_2009}. These images were identified and extracted using a custom opencv-based segmentation algorithm consisting of image binarization, denoising, and glyph segment isolation. To ensure the extracted text components are correctly grouped and arranged in correct order, spatial clustering methods were adopted to ensure complete, legal Manchu text are reconstructed from the extracted glyph segments. After extraction, these images were subjected to image processing routines to approximate the visual style and format of the synthetic training data, ensuring consistency in evaluation.

\begin{center}
	\includegraphics[width=0.2\columnwidth,clip,trim=0.1cm 0.1cm 0.1cm 0.1cm]{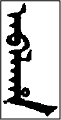}
	\captionof{figure}{Original hand-written Manchu word from \textit{Neige Cangben Manwen Laodang}.}
	\label{fig:woodblock_word}
\end{center}

\begin{center}
	\includegraphics[width=\columnwidth,clip,trim=0.1cm 0.1cm 8cm 0.1cm]{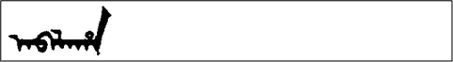}
	\captionof{figure}{Hand-written Manchu word after processing for evaluation.}
	\label{fig:woodblock_after_processing}
\end{center}

\section{Methodology}
\subsection{Model preparation}

We selected three open-source VLMs with varying parameter sizes and architectures to assess the effect of model scale and design on OCR accuracy: Qwen2.5-VL-3B, Qwen2.5-VL-7B (both from Alibaba) and LLaMA-3.2-11B (Meta). This selection was motivated by several factors: (1) Parameter diversity spanning 3B to 11B parameters allows systematic investigation of scale effects on OCR performance, as recent studies indicate that while larger models generally provide superior OCR capabilities~\citep[pp.~1]{large_ocr_scaling}, well-optimized smaller models can achieve competitive results through architectural improvements~\citep[pp.~1]{xmodel_vlm}; (2) Both Qwen2.5-VL and LLaMA-3.2, despite sharing similar multimodal capabilities~\citep{qwen25vl,llama3_herd}, offer different parameter scales and training optimizations that provide insights into low-resource script recognition; (3) Both model families have demonstrated state-of-the-art performance on vision-language benchmarks, with Qwen2.5-VL consistently achieving top scores on OCR tasks~\citep{qwen25vl} and LLaMA-3.2-11B excelling in document VQA and high-resolution image processing up to 1120×1120 pixels~\citep{llama3_herd}.

Each model was fine-tuned using the \textsl{FastVisionModel} wrapper from the \textsl{Unsloth} framework~\citep{unsloth2023}, which supports efficient parameter-efficient fine-tuning (PEFT) of both visual and language components. Our training objective was to generate structured OCR output containing both the Manchu script and its romanized transliteration. To achieve this, we designed a consistent instruction format and applied identical hyperparameters across all models to ensure fair comparison. The detailed prompt design and training configuration are described in the following sections.

\subsection{Prompt Design}
Prompt design played a critical role in adapting general-purpose VLMs to the OCR task. We designed a structured prompt to guide the model to generate outputs in a dual-script format. Each input image was paired with the following prompt:

\begin{quote}
	\raggedright%
	\small\ttfamily You are an expert OCR system for Manchu script. Extract the text from the provided image with perfect accuracy. Format your answer exactly as follows: first line with `Manchu:' followed by the Manchu script, then a new line with `Roman:' followed by the romanized transliteration.
\end{quote}

This instruction was applied uniformly across all training examples without distinction between system and user prompts, treating it as a single unified instruction for the OCR task. The target output was formatted to include both the Unicode-based Manchu script (rendered using Mongolian Unicode due to script similarity) and its corresponding Roman transliteration, encouraging the model to learn both visual decoding and phonetic transduction. 

For example, given an image containing the Manchu word for ``behind'' or ``after, later'', the expected output format would be:

\begin{quote}
	\raggedright%
	\small\ttfamily
	Manchu:\manchu{ᠠᠮᠠᠯᠠ}\\
	Roman:AMALA\\
\end{quote}

The prompt was held constant across all models to control for variation in instruction interpretation. We did not conduct prompt ablation or variation studies in this work, focusing instead on evaluating the effect of model architecture and size under a fixed instruction format.

\subsection{Training}

\subsubsection{Hyperparameter Configuration}

To optimize training efficiency while maintaining model performance, we employed differentiated epoch schedules based on model size: Qwen2.5-VL-3B for 15 epochs, Qwen2.5-VL-7B for 10 epochs, and LLaMA-3.2-11B for 5 epochs. This inverse relationship between model size and training epochs is motivated by empirical findings that larger models are more sample efficient and converge faster than smaller counterparts, while training beyond 4 epochs yields diminishing returns and can lead to performance degradation~\citep{tinyllama_efficiency}. The graduated schedule allows smaller models to benefit from extended training while preventing overfitting in larger models, optimizing computational efficiency across different parameter scales. All models used the AdamW optimizer with an 8-bit precision variant (\texttt{paged\_adamw\_8bit}) to reduce memory consumption. The learning rate was set to $1 \times 10^{-4}$ with cosine learning rate scheduling and warm-up over the first 50 steps.

Training was performed with a per-device batch size of 4 and gradient accumulation over 2 steps to effectively simulate a batch size of 8. Mixed-precision training was enabled via bfloat16 (bf16) to accelerate training and reduce memory usage. We used weight decay of 0.05.

All models were fine-tuned using the \textsl{Unsloth} training framework's \textsl{SFTTrainer}, which wraps HuggingFace's \texttt{transformers} Trainer with additional support for multimodal instruction tuning. Training was conducted on an NVIDIA RTX A6000 GPU with Ubuntu 24.04 LTS; detailed hardware and software specifications are provided in Appendix~\ref{app:hardware_specs}. We employed LoRA-based parameter-efficient fine-tuning (PEFT) for both vision and language components, as LoRA enables faster training and significantly reduced memory requirements compared to full fine-tuning while maintaining competitive performance~\citep{hu2022lora}. The LoRA configuration used a rank of 64, $\alpha = 64$, and no dropout. These hyperparameters were selected through preliminary experiments to balance training efficiency and model performance, following common practices in multimodal fine-tuning literature.

\columntable{ll}{
\toprule
\textbf{Hyperparameter} & \textbf{Value} \\
\midrule
Number of epochs & 15 (3B), 10 (7B), 5 (11B) \\
Optimizer & AdamW (8-bit) \\
Learning rate & $1 \times 10^{-4}$ \\
LR scheduler & Cosine with warmup \\
Warmup steps & 50 \\
Per-device batch size & 4 \\
Gradient accumulation steps & 2 \\
Effective batch size & 8 \\
Mixed precision & BFloat16 (bf16) \\
Weight decay & 0.05 \\
Random seed & 3407 \\
LoRA rank & 64 \\
LoRA alpha & 64 \\
LoRA dropout & 0.0 \\
\bottomrule
}{Training hyperparameters for VLM fine-tuning\label{tab:hyperparams}}

\subsection{Evaluation Metrics}

To assess OCR performance, we employed four primary evaluation metrics that capture different aspects of model accuracy and practical usability. All metrics were computed for each sample and aggregated over the evaluation set:

\textbf{Character Error Rate (CER)}: We calculated CER using the Levenshtein distance (edit distance)~\citep{levenshtein1965distance} between the predicted Manchu string and the reference, normalized by the total number of characters in the ground truth. This standard OCR metric captures insertion, deletion, and substitution errors at the character level.

\textbf{Word Accuracy}: We measured exact-match accuracy at the word level, computing the percentage of test samples where the model's predicted Manchu script exactly matches the ground truth. This metric provides a direct assessment of the model's ability to correctly recognize complete Manchu words without any character-level errors.

\textbf{F1 Score}: We implemented a character-level F1 score based on sequence matching to balance precision and recall. We identified matching character blocks between predicted and ground truth sequences to calculate true positives. Precision was computed as the ratio of true positives to the total predicted characters, while recall was the ratio of true positives to the total ground truth characters. The F1 score was then calculated as the harmonic mean of precision and recall:

\begin{equation}
F1 = 2 \times \frac{\text{precision} \times \text{recall}}{\text{precision} + \text{recall}}
\end{equation}

For illustration, consider these scenarios:

\textbf{Under-generation (Deletion):}
\begin{itemize}[label={}]
	\item Ground truth: \texttt{\manchu{ᠠᠮᠠᠯᠠ}} (5 characters)
	\item Predicted: \texttt{\manchu{ᠠᠮᠠ}} (3 characters)
	\item True positives: 3 (matching block \manchu{ᠠᠮᠠ})
	\item Precision: 3/3 = 1.0
	\item Recall: 3/5 = 0.6
	\item F1 Score: 0.75
\end{itemize}

\textbf{Over-generation (Insertion):}
\begin{itemize}[label={}]
	\item Ground truth: \texttt{\manchu{ᠠᠮᠠᠯᠠ}} (5 characters)
	\item Predicted: \texttt{\manchu{ᠠᠮᠠᠯᠠᠠᠮᡳᠨ}} (8 characters)
	\item True positives: 5 (matching block \\ \manchu{ᠠᠮᠠᠯᠠ})
	\item Precision: 5/8 = 0.625
	\item Recall: 5/5 = 1.0
	\item F1 Score: 0.77
\end{itemize}

This metric effectively captures the model's ability to correctly identify characters while penalizing both over-generation and under-generation errors.

\textbf{Inference Time}: We measured the average time required for each model to process a single image and generate the OCR output, including both the Manchu script and its romanized transliteration. This metric is crucial for assessing the practical feasibility of deploying the OCR system in real-world applications, where processing speed directly impacts user experience and system scalability.

\section{Results}

\subsection{Model Selection and Training Stability}

\begin{figure*}[ht!]
    \centering
    \begin{subfigure}[b]{0.49\textwidth}
        \includegraphics[width=\textwidth]{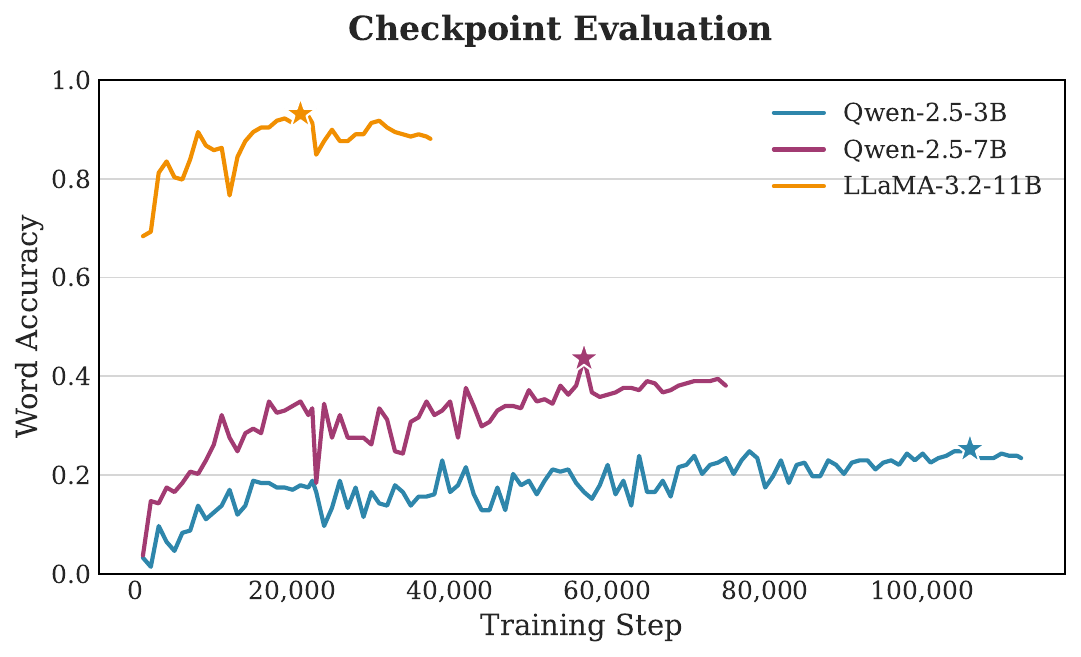}
        \caption{Checkpoint evaluation trajectories for word accuracy.}
        \label{fig:checkpoint_trends}
    \end{subfigure}
    \hfill 
    \begin{subfigure}[b]{0.49\textwidth}
        \includegraphics[width=\textwidth]{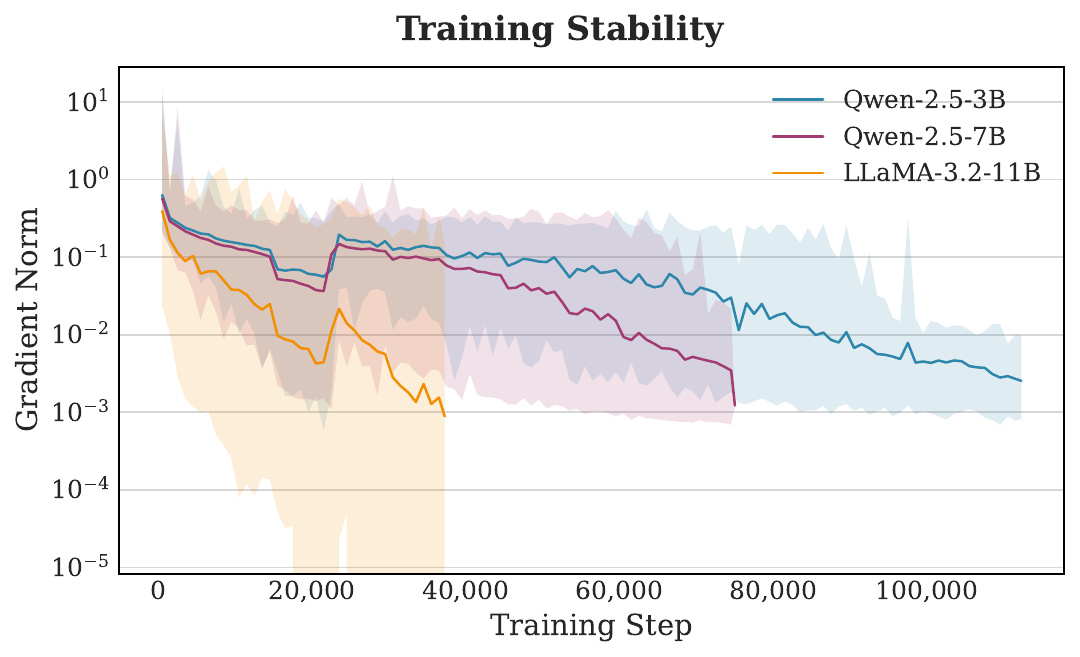}
        \caption{Gradient norm trajectories during training.}
        \label{fig:training_performance}
    \end{subfigure}
    \caption{Training dynamics and checkpoint selection}
    \label{fig:training_and_checkpoints}
\end{figure*}

We saved model checkpoints at every 1,000 training steps to monitor training progression comprehensively. Following training completion, we evaluated all checkpoints on our real-world test dataset and selected those achieving the highest word accuracy performance. This post-training evaluation strategy ensures optimal model performance while avoiding dependence on intermediate validation metrics that may not align with final task objectives.

Figure~\ref{fig:checkpoint_trends} reveals distinct convergence patterns across model architectures. The LLaMA-3.2-11B model demonstrated superior efficiency, achieving peak word accuracy after only 21,000 training steps, while the Qwen variants required substantially longer training periods to reach their optimal performance points. Specifically, the selected optimal checkpoints were: Qwen2.5-VL-3B at step 106,000, Qwen2.5-VL-7B at step 57,000, and LLaMA-3.2-11B at step 21,000. These convergence differences reflect underlying architectural efficiencies, with the larger LLaMA model reaching optimal accuracy approximately three to five times faster than the smaller Qwen variants.

To further assess training stability, we monitored gradient norm throughout the training process as a key indicator of optimization health. Gradient norm serves as a critical diagnostic metric for understanding training dynamics: excessive values may signal unstable optimization or poor hyperparameter choices, while consistently low values indicate smooth convergence and well-conditioned parameter updates. 

Figure~\ref{fig:training_performance} illustrates the gradient norm trajectories throughout training for all three models. The analysis reveals markedly different optimization behaviors across architectures. LLaMA-3.2-11B demonstrates the most stable gradient dynamics, with smooth convergence to low gradient norms, indicating well-conditioned parameter updates and effective optimization. The Qwen variants, particularly the 3B model, exhibit more volatile gradient trajectories with higher final gradient norms, suggesting less stable optimization landscapes that may require different hyperparameter configurations or longer training periods to achieve optimal convergence.

These gradient norm patterns align with the observed performance differences, where models with more stable gradient dynamics (LLaMA-3.2-11B) achieved superior OCR accuracy. The correlation between gradient stability and final performance underscores the importance of monitoring optimization health beyond traditional loss metrics when fine-tuning vision-language models for specialized tasks.

\subsection{Overall Performance}

Evaluation was conducted on 1,000 randomly selected samples from the synthetic validation dataset comprising diverse glyph sequences, as well as the real-world handwritten test dataset from the \textsl{Neige Cangben Manwen Laodang}. The evaluation metrics included Word Accuracy, CER, F1 score, and Inference Time for Manchu script recognition. Table~\ref{tab:performance_comparison} summarizes performance across both datasets.

\begin{table*}[b]
\centering
\begin{tabular}{lcccc}
	\toprule
	\textbf{Model} & \textbf{Word Accuracy (\%)} & \textbf{CER} & \textbf{F1 Score} & \textbf{Inference Time (s)} \\
	\midrule
	\multicolumn{5}{c}{\textit{Validation Dataset (1,000 synthetic samples)}} \\
	\midrule
	LLaMA-3.2-11B & \textbf{98.3} & \textbf{0.0024} & \textbf{0.998} & 14.7 \\
	Qwen2.5-VL-7B & 87.5 & 0.0264 & 0.978 & 1.3 \\
	Qwen2.5-VL-3B & 84.4 & 0.0329 & 0.973 & \textbf{1.2} \\
	\midrule
	\multicolumn{5}{c}{\textit{Test Dataset (218 handwritten samples)}} \\
	\midrule
	LLaMA-3.2-11B & \textbf{93.1} & \textbf{0.0219} & \textbf{0.983} & 8.9 \\
	Qwen2.5-VL-7B & 43.1 & 0.254 & 0.789 & 0.9 \\
	Qwen2.5-VL-3B & 23.9 & 0.368 & 0.709 & \textbf{0.7} \\
	\bottomrule
\end{tabular}
\caption{Performance comparison across validation and test datasets}
\label{tab:performance_comparison}
\end{table*}

On the synthetic validation dataset, LLaMA-3.2-11B achieved exceptional performance with 98.3\% word accuracy, 0.998 F1 score, and remarkably low CER of 0.0024. The Qwen variants demonstrated competitive but lower performance: Qwen2.5-VL-7B reached 87.5\% accuracy with 0.978 F1 score, while Qwen2.5-VL-3B achieved 84.4\% accuracy with 0.973 F1 score. However, the Qwen models offered significant speed advantages, processing samples in 1.2-1.3 seconds compared to LLaMA's 14.7 seconds per sample.

The performance landscape changes dramatically on real-world test data. LLaMA-3.2-11B maintained robust performance with 93.1\% word accuracy and 0.983 F1 score, representing only a modest 5.2 percentage point accuracy decline from validation. In stark contrast, the Qwen models suffered severe performance degradation: Qwen2.5-VL-7B dropped to 43.1\% accuracy (44.4 point decline) and Qwen2.5-VL-3B fell to 23.9\% accuracy (60.5 point decline). These results reveal fundamental differences in generalization capability, with LLaMA's superior architecture enabling effective transfer from synthetic training data to real historical documents, while the Qwen models exhibit substantial overfitting to synthetic patterns.

\subsection{Comparison with Traditional OCR Approaches}

\begin{figure*}
	\centering
	\includegraphics[width=1\textwidth]{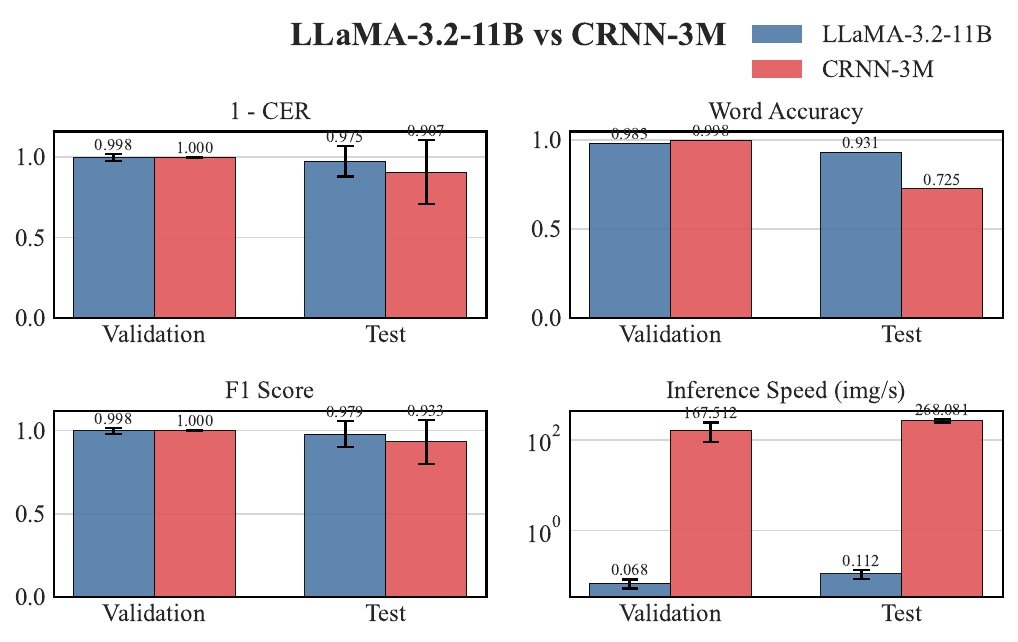}
	\caption{Performance comparison between CRNN and LLaMA-3.2-11B models}
	\label{fig:crnn_vs_llama_comparison}
\end{figure*}

To contextualize the VLM performance, we conducted a comprehensive comparative evaluation against a traditional CRNN-based OCR system, representing the established approach for sequence recognition tasks in computer vision. The CRNN architecture combines convolutional neural networks for feature extraction with recurrent neural networks for sequence modeling, followed by connectionist temporal classification (CTC) decoding for alignment-free recognition. Detailed experimental configurations for the CRNN baseline are provided in Appendix~\ref{app:crnn_setup}. Following the same rigorous checkpoint selection methodology as the VLM experiments, we evaluated all training checkpoints and identified epoch 48 as the optimal checkpoint for the CRNN model based on validation performance.

On the synthetic validation set, the CRNN model demonstrated exceptional specialization for Manchu script recognition, achieving 99.8\% word accuracy, 0.9999 F1 score, and remarkably low CER of 0.0001. This performance substantially outperformed the LLaMA model on synthetic data, highlighting the effectiveness of task-specific architectural design when training and test distributions are well-matched. The CRNN's superior performance on controlled synthetic data reflects its optimized feature extraction pipeline, specifically tuned for the visual patterns present in rendered Manchu script.

However, this advantage reversed dramatically when evaluated on real-world test data. On the 218 handwritten samples, CRNN performance deteriorated significantly to 72.5\% word accuracy, 0.933 F1 score, and 0.093 CER, representing a substantial 27.3 percentage point accuracy decline.
Interestingly, while the CRNN model still outperformed both Qwen variants on real-world data (72.5\% vs 43.1\% and 23.9\%), it remained substantially inferior to LLaMA-3.2-11B's robust 93.1\% accuracy. This comparison reveals that while architectural specialization can achieve superior performance on matched distributions, the multimodal understanding and pre-trained representations of large vision-language models provide crucial advantages for cross-domain generalization. Figure~\ref{fig:crnn_vs_llama_comparison} illustrates this performance comparison, highlighting the superior generalization capability of the vision-language model approach and the fundamental trade-offs between specialized optimization and robust adaptability.

\subsection{Comparison with Commercial Models}
To assess the practical viability of commercial OCR solutions for Manchu, we fine-tuned a closed-domain model (\textsl{GPT-4.1-2025-04-14}) on a subset of our training data. Due to financial constraints, we trained on only one-fourth of our validation dataset (15,000 samples) randomly selected from the original 60,000-sample training set. The model was trained for a single epoch with a batch size of 9 and a learning rate multiplier of 2.0, consuming 5,696,401 tokens during training.

Given the GPT-4.1 pricing structure of \$25.00 per million training tokens (as of May 31, 2025), this limited experiment incurred a cost of approximately \$142.41. A complete training regimen matching our LLaMA-3.2-11B configuration (60,000 samples × 5 epochs) would require approximately 113,928,020 tokens, resulting in an estimated cost of \$2,848.20.

The GPT-4.1 model's performance on 218 test samples was remarkably poor, achieving 0\% word accuracy for both Manchu script and Roman transliteration, CER values exceeding 1.0, and F1 scores below 0.22. The inference time averaged 2.567 seconds per sample, which was also substantially slower than our open-source VLM implementations. Note that this inference time is significantly affected by network latency as the GPT-4.1 model was accessed via REST API.

\section{Findings and Discussion}

\subsection{Error Analysis }

Detailed examination of successful predictions demonstrates the models' capacity for complex sequence recognition. The longest correctly predicted string by LLaMA-3.2-11B spans 13 characters:

\begin{quote}
	\raggedright%
	\textbf{Manchu:} {\small\manchu{ᡶᡝᡵᡤᡠᠸᡝᠪᡠᡵᠠᡴᡡᠩᡤᡝ}} \par
 	\textbf{Roman:} FERGUWEBURAKŪNGGE
\end{quote}

This example illustrates the model's ability to maintain accuracy across extended character sequences, including complex consonant clusters and vowel harmonies characteristic of Manchu phonology.

Nevertheless, our analysis reveals systematic error patterns stemming from morphological similarity between Manchu glyphs. Table~\ref{tab:top_errors} shows that all models exhibit systematic vowel confusion, with LLaMA-3.2-11B demonstrating the highest error concentration (55.9\% on test data), indicating more systematic rather than random failure patterns. The universal difficulty with \manchu{ᠠ} (A) across all models—appearing as the top error in 5 out of 6 cases—reflects the inherent morphological complexity of Manchu script. Most failures occur with characters sharing similar or even identical stroke patterns and positional variants, particularly among vowels (\manchu{ᠠ} (A), \manchu{ᡝ} (E), \manchu{ᡳ} (I)). In some cases, these glyphs are visually indistinguishable in isolation and can only be correctly identified by considering the preceding and succeeding characters, as well as subtle diacritical marks, stroke orientations, and their position within the syllable (initial, medial, or final). 
\begin{figure*}
	\centering
	\includegraphics[width=\textwidth]{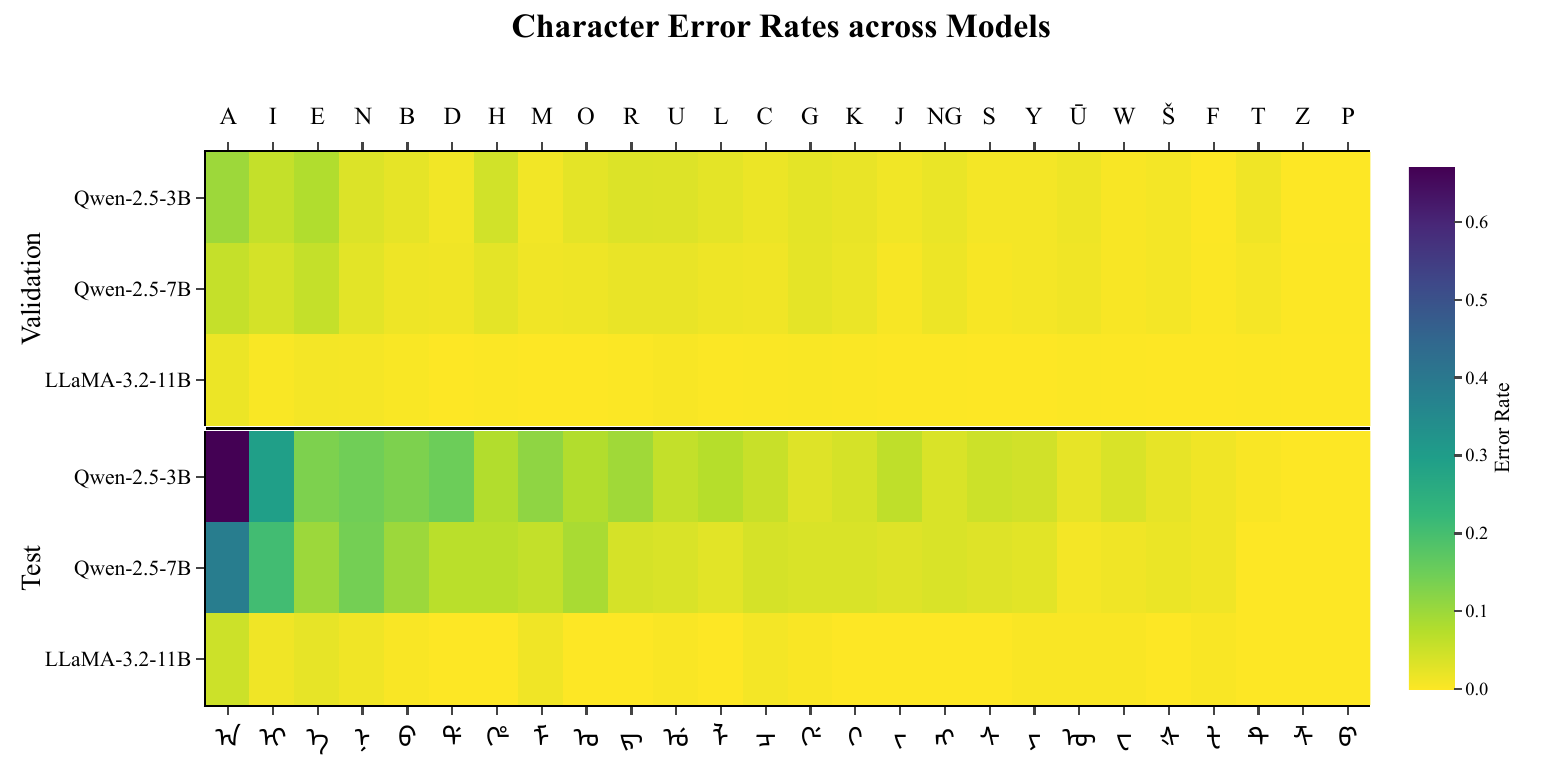}
	\caption{Top misrecognized characters in Manchu predictions}
	\label{fig:error_characters_heatmap}
\end{figure*}

\begin{table*}[t]
\centering
\begin{tabular}{lcc}
	\toprule
	\textbf{Model} & \textbf{Top 3 Characters} & \textbf{Concentration} \\
	\midrule
	\multicolumn{3}{c}{\textit{Validation Dataset (1,000 synthetic samples)}} \\
	\midrule
	LLaMA-3.2-11B & \manchu{ᠠ} (A) (25.7\%), \manchu{ᡝ} (E) (12.9\%), \manchu{ᠨ} (N) (11.4\%) & 50.0\% \\
	Qwen2.5-VL-7B & \manchu{ᡝ} (E) (12.8\%), \manchu{ᠠ} (A) (12.6\%), \manchu{ᡳ} (I) (9.1\%) & 34.5\% \\
	Qwen2.5-VL-3B & \manchu{ᠠ} (A) (15.6\%), \manchu{ᡝ} (E) (12.3\%), \manchu{ᡳ} (I) (9.2\%) & 37.2\% \\
	\midrule
	\multicolumn{3}{c}{\textit{Test Dataset (218 handwritten samples)}} \\
	\midrule
	LLaMA-3.2-11B & \manchu{ᠠ} (A) (32.4\%), \manchu{ᡝ} (E) (14.7\%), \manchu{ᠮ} (M) (8.8\%) & 55.9\% \\
	Qwen2.5-VL-7B & \manchu{ᠠ} (A) (23.6\%), \manchu{ᡳ} (I) (12.6\%), \manchu{ᠨ} (N) (8.7\%) & 44.9\% \\
	Qwen2.5-VL-3B & \manchu{ᠠ} (A) (27.2\%), \manchu{ᡳ} (I) (11.9\%), \manchu{ᡩ} (D) (6.2\%) & 45.3\% \\
	\bottomrule
\end{tabular}
\caption{Character-level error analysis on test and validation data}
\label{tab:top_errors}
\end{table*}

To address the issue, we are now developing a new dataset that is composed of scanned Manchu texts. This dataset, similar to the real-world test set employed in this study, will be used to train a new model that is specifically designed to handle the morphological similarity between Manchu glyphs. It is expected, by exposing the model to more diverse visual patterns, the model will be able to generalize better to the real-world data.



\subsection{Performance Comparison between VLM and CRNN}

The substantial generalization advantage of VLMs highlights the contextual understanding capabilities inherent in vision-language models. While traditional CNN-based approaches excel at pattern recognition within constrained domains, VLMs leverage linguistic knowledge to maintain coherence even when encountering visually ambiguous or degraded characters. The dramatic performance gap between validation and test sets for CRNN (99.8\% vs 72.5\%) compared to the stable performance of LLaMA (98.3\% vs 93.1\%) underscores the fundamental advantage of multimodal language models for low-resource OCR applications.

The CRNN's severe overfitting to synthetic training patterns demonstrates the limitations of traditional approaches when faced with real-world document variability. In contrast, the robust cross-domain performance of VLMs suggests that their pre-trained linguistic representations provide valuable inductive biases for handling script variations and degradation typical in historical documents.

\subsection{Open-Source vs. Commercial Model Paradigms}

Our comparative evaluation reveals several critical disadvantages of commercial closed-domain AI for low-resource language applications. First, the prohibitive cost structure fundamentally limits the extensive fine-tuning required for specialized domains. While closed-domain models derive their strength from versatility across common languages and tasks, low-resource languages like Manchu are typically absent from their pre-training data, necessitating substantial domain adaptation. However, the financial barrier makes extensive fine-tuning economically unfeasible. Our limited single-epoch training already cost \$142.41, while a comprehensive training regimen would exceed \$2,800.

Second, commercial models impose architectural constraints that hinder optimization for specialized tasks. Unlike open-source alternatives where researchers can modify model architecture, attention mechanisms, and training procedures to suit specific linguistic features, closed-domain systems offer limited customization options. This inflexibility is particularly problematic for scripts like Manchu, which require specialized tokenization, character-level understanding, and culturally-specific linguistic patterns.

Third, the dependency on API access introduces latency, reliability, and data privacy concerns. Researchers cannot guarantee consistent performance, control inference speed, or ensure data security when processing sensitive historical documents through external services. Additionally, commercial models may be discontinued, pricing may change arbitrarily, or access may be restricted, creating sustainability risks for long-term research projects. Furthermore, geographical restrictions limit access to state-of-the-art commercial models in many countries, creating inequitable barriers for international researchers working on endangered language preservation—a field that inherently requires global collaboration and diverse scholarly perspectives.

In contrast, open-source VLMs like LLaMA-3.2-11B can be executed on accessible infrastructure such as Google Colab, achieving superior performance at a fraction of the cost while offering full control over the training process, model architecture, and deployment environment. This fundamental advantage makes open-source approaches essential for endangered and low-resource languages, where research budgets are constrained but linguistic preservation is critical.

\section{Conclusion}
This study demonstrates that state-of-the-art optical character recognition (OCR) for low-resource and endangered languages is now achievable using open-source vision-language models (VLMs) that can be fine-tuned and deployed locally with modest computational resources. By leveraging synthetic data and parameter-efficient training, our LLaMA-3.2-11B-based system achieves a character error rate (CER) of 0.0024 and 98.3\% word-level accuracy on synthetic benchmarks. It also demonstrates promising potential in generalizing to real-world Manchu handwritten texts.

A key finding of this work is that high-quality synthetic datasets can be used to train OCR systems that perform well on real historical materials. This is especially crucial for digital humanities researchers, who often lack the resources or access to build large annotated datasets from real-world sources. Our results show that with careful preprocessing and model design, synthetic data can bridge the gap between resource constraints and the need for accurate, scalable OCR solutions.

Equally important, our approach can be run entirely on a single workstation or in accessible environments such as Google Colab, making advanced OCR technology available to humanities scholars, archivists, and independent researchers regardless of institutional resources. This local, cost-effective, and fully open-source solution addresses critical barriers in digital humanities, where data privacy, sustainability, and reproducibility are paramount.


This Manchu OCR system will significantly lower the technical barrier for historians and linguists working with Manchu archival materials. The complete implementation, including model weights, training scripts, and evaluation code, is publicly available at \url{https://github.com/mic7ch1/ManchuAI-OCR} to facilitate reproducibility and enable further research in low-resource language OCR. We hope this work will inspire further collaboration and innovation at the intersection of artificial intelligence and the digital humanities, contributing to the preservation and revitalization of the world's linguistic heritage.

\bibliographystyle{unsrtnat}
\bibliography{reference}
\newpage

\appendix
\part*{Appendices}

\section{Hardware and Software Specifications}
\label{app:hardware_specs}

\subsection{Computing Infrastructure}

\columntable{ll}{
\toprule
\textbf{Component} & \textbf{Specification} \\
\midrule
GPU & NVIDIA RTX A6000 Ada (48GB VRAM) \\
CPU & Intel i9-13900KS (32 cores) \\
RAM & 188GB DDR4 \\
Storage & 3.6TB NVMe SSD \\
\bottomrule
}{Hardware specifications for model training.\label{tab:hardware_specs}}

\subsection{Software Environment}

\columntable{ll}{
\toprule
\textbf{Component} & \textbf{Version} \\
\midrule
Operating System & Ubuntu 24.04 LTS \\
CUDA & 12.4 \\
Python & 3.10 \\
PyTorch & 2.6 \\
Transformers & 4.50 \\
Datasets & 3.3 \\
Unsloth & 2025.4.7 \\
\bottomrule
}{Software environment specifications.\label{tab:software_specs}}

\section{CRNN Baseline Configuration}
\label{app:crnn_setup}

\subsection{Model Architecture}

The CRNN baseline implements a deep convolutional recurrent architecture optimized for sequence recognition tasks. The model comprises three main components: a 9-layer convolutional feature extractor, a 4-layer bidirectional LSTM sequence modeler, and a linear classification layer with CTC decoding.

\columntable{ll}{
\toprule
\textbf{Component} & \textbf{Specification} \\
\midrule
\multicolumn{2}{l}{\textit{CNN Feature Extractor}} \\
Layers & 9 convolutional layers \\
Channel progression & 3 → 64 → 128 → 256 → 512 \\
Input size & 64 × 480 pixels \\
Output features & 512-dimensional per position \\
\midrule
\multicolumn{2}{l}{\textit{LSTM Sequence Modeler}} \\
Architecture & 4-layer bidirectional LSTM \\
Hidden units & 512 per direction (1024 total) \\
Dropout rate & 10\% between layers \\
\midrule
\multicolumn{2}{l}{\textit{Output Layer}} \\
Type & Linear + CTC decoding \\
Function & Character class probabilities \\
\bottomrule
}{CRNN model architecture overview.\label{tab:crnn_arch}}

\textbf{Convolutional Feature Extractor:} The CNN backbone processes 64×480 pixel inputs through nine convolutional layers with progressive channel expansion (3→64→128→256→512). Each convolutional block includes batch normalization, ReLU activation, and light dropout (5\% for CNN layers). Strategic max pooling operations reduce spatial dimensions while preserving sequence length: initial pooling reduces height to 32×240, then 16×120, followed by asymmetric pooling operations (2×1 kernel) that compress height to 4×120 while maintaining width for sequence modeling. The final layers use adaptive average pooling to produce 512-dimensional features per sequence position.

\textbf{Recurrent Sequence Modeling:} Feature maps are reshaped and fed into a 4-layer bidirectional LSTM with 512 hidden units per direction, producing 1024-dimensional contextual representations. The deep LSTM architecture with 10\% dropout between layers enables robust sequence modeling for variable-length Manchu word recognition.

\textbf{Classification and Decoding:} A final linear layer with dropout projects the bidirectional LSTM outputs to character class probabilities, followed by CTC decoding to handle alignment-free sequence prediction without explicit character segmentation.

\subsection{Training Configuration}

\columntable{ll}{
\toprule
\textbf{Parameter} & \textbf{Value} \\
\midrule
Dataset & 60,000 validation samples \\
Optimizer & AdamW \\
Initial learning rate & 1e-3 \\
Weight decay & 0.01 \\
LR scheduler & CosineAnnealingWarmRestarts \\
Input image size & 64 × 480 pixels \\
Hidden size & 256 \\
Dropout & 0.1 \\
Normalization & [0, 1] \\
Data augmentation & Random rotation (±5°), perspective \\
                  & transformation, Gaussian noise \\
Training epochs & 100 \\
Warmup epochs & 5 \\
Mixed precision & True \\
Gradient clipping & Max norm 1.0 \\
Batch size & 16 \\
\bottomrule
}{CRNN training configuration.\label{tab:crnn_training}}

\subsection{Evaluation Protocol}

The evaluation protocol mirrored that of the VLM experiments to ensure a consistent comparison. The validation set consisted of the full 15,000 synthetic images. Due to the CRNN model's high inference speed, we evaluated it on the entire set, in contrast to the 1,000-sample subset used for reporting VLM performance. The real-world test set was identical, containing 218 handwritten Manchu words. All metrics (word accuracy, character error rate (CER), and F1 score) were computed using the same implementations as in the VLM assessment.

\end{document}